\documentclass[12pt]{article}
\usepackage[xcolor={divpdf,grey},authormarkup=none]{changes} 
\usepackage{changes} 
\definechangesauthor[name={Mayita},color=red]{MGL}

\usepackage{amsmath}%
\usepackage{amsfonts}%
\usepackage{amssymb}%
\usepackage{graphicx}
\usepackage{pictex}
\usepackage{color}
\usepackage{setspace}

\usepackage{longtable}
\usepackage{caption}
\usepackage{subcaption}
\usepackage{tabularx}
\usepackage{geometry}
\usepackage{makecell}

\usepackage{authblk}

\newcommand {\bc}{\begin{center}}
	\newcommand {\ec}{\end{center}}
\usepackage{verbatim}
\usepackage{mathrsfs}
\usepackage{latexsym}
\newcommand{\dado}{\,|\,}

\newcommand{\prob}{\text{Pr}}

\newcommand{\rea}{\mathbb{R}}

\newcommand{\logn}{\textrm{ln}}
\newcommand{\expn}{\textrm{exp}}

\newcommand{\dis}{\mbox{dist}}

\newcommand*{\qeda}{\hfill\ensuremath{\square}}

\newcommand{\ve}{\mathbf}

\newcommand{\eps}{\epsilon}
\newcommand{\beq}{\begin{equation}}
\newcommand{\enq}{\end{equation}}
\newcommand{\beqa}{\begin{eqnarray}}
\newcommand{\enqa}{\end{eqnarray}}
\newcommand{\neno}{\newline\noindent}

\newcommand{\cald}{{\cal D}}

\newcommand{\calm}{{\cal M}}

\newcommand{\calx}{{\cal X}}

\newcommand{\ogd}{\mbox{O}}
\newcommand{\och}{\mbox{o}_{\Pr}}

\newcommand{\real}{\mathbb{R}}

\newtheorem{theorem}{Theorem}

\newtheorem{definition}{Definition}

\newtheorem{proposition}{Proposition}

\title{A bagging and importance sampling approach to Support Vector Machines}

	\author[1]{\small B\'arcenas R.\thanks{roberto.barcenas@cimat.mx}}
	\author[2]{\small G\'onzalez--Lima M. D.\thanks{maria.gonzalezl@unimilitar.edu.co}}
	\author[3]{\small \enskip Quiroz A. J.\thanks{aj.quiroz1079@uniandes.edu.co}}
	
	\affil[1]{\footnotesize Centro de Investigaci\'on en Matem\'aticas, Guanajuato, M\'exico}
	\affil[2]{\footnotesize Universidad Militar Nueva Granada, Bogot\'a, Colombia}
	\affil[3]{\footnotesize Dpto. de Matem\'aticas, Universidad de los Andes, Bogot\'a, Colombia}

\date{}

\begin{document}
	
	
	\maketitle
	
	

	
\begin{abstract}
	
	An importance sampling and bagging approach to solving the support vector 
	machine (SVM) problem in the context of large databases is presented and evaluated. 
	Our algorithm builds on the nearest neighbors ideas presented in \cite{Camelo}. 
	As in that reference, the goal of the present proposal is to achieve a faster 
	solution of the SVM problem without a significance loss in the prediction error.
	The performance of the methodology is evaluated in benchmark examples 
	and  theoretical aspects of subsample methods are discussed.

	
\end{abstract}

\section{Introduction}

In the context of pattern recognition, classification methods focus on learning 
the relationship between a set of feature variables and a target ``class 
variable'' of interest. From a set of training data points with known
associated class labels, a classifier is adjusted, to be
used on unlabeled test instances. The diversity of problems that can be 
addressed by classification algorithms is significant and covers many domains
of application (see, for instance, \cite{Cart} or \cite{Hastie}). More 
formally, in the setting of \emph{supervised learning}, examples are given 
that consist of pairs, $(X_{i},y_{i})$, $i \leq n$, where $X_{i}$ is the 
$d-$dimensional ``feature'' or ``covariate'' 
vector and $y_{i}$ is the corresponding  ``class'' 
or  ``category'', that lives in some finite set ${\cal C}$. Frequently, as in the present work,
${\cal C}$ is assumed to consist only of two classes, labeled $1$ and $-1$. 

In recent decades, diverse non-linear procedures have been considered
for the classification problem. These techniques have been reported to have a superior performance,
in many important problems, with respect to their classical linear counterparts. Among the main non-
linear methods, one should mention Classification and Regression Trees (CART) and its general version, 
Random Forest, Neural Networks, Nearest Neighbor Classifiers, Support Vector  Machines (SVM) and 
probabilistic methods. For detailed accounts of
these methodologies, the reader can consult \cite{Cart}, \cite{Breiman}, \cite{Breiman2}, \cite{Duda}, \cite{Devroye}, \cite{Hastie} and \cite{CrisShaw}.

Support Vector Machines is a widely used approach in classification, introduced 
by Boser, Guyon and Vapnik \cite{Boser}, combining ideas from linear 
classification methods, Optimization and Reproducing Kernel Hilbert Spaces. 
(See also \cite{Cortes} and \cite{CrisShaw} for details). The 
success of SVM in terms of classification error in a variety of 
contexts, can be explained, in part, by their flexibility. The method uses a kernel function as inner 
product of ``new'' feature variables that come from a transformation of the original 
covariates. Another reason for the success of this methodology is
the effort that has been invested in developing efficient methods
of solution, some of which, we will discuss below.

With the notation given above for the training data, the {\it soft margin $L^1$ classifier} SVM problem is stated as
\begin{equation}\label{sml1prim}
\begin{array}{ll}
\underset{w,b,\xi}{\mbox{minimize}} & \frac{1}{2} \| w\|^2 +C\sum_{i=1}^n\xi_i \quad \\
{\rm subject}\; {\rm to} & y_i (w^t \phi(X_i) + b) \geq 1-\xi_i,  \;\; \forall \; i = 1,...,n,\\ 
&  \xi_i \geq 0
\end{array}
\end{equation}

\noindent where $\phi(\cdot)$ is the transformation of the feature vector into a 
higher dimensional space induced by the use of a kernel function, $C$ is a 
positive constant that expresses the cost of loosing separation margin or 
misclassification. The $\xi_i$ are the slack variables. The $L^2$ version of the soft margin classifier is obtained by
replacing the objective function by \[
\frac{1}{2} \| w\|^2 +C\sum_{i=1}^n\xi^2_i  
\]
in (\ref{sml1prim}). The dual problem corresponding to (\ref{sml1prim}) can be written as
\begin{equation}\label{sml1dual}
\begin{array}{ll}
\underset{\alpha}{\mbox{maximize}} & \sum_{i=1}^n\alpha_i -
\frac{1}{2} \sum_{i,j=1}^n y_iy_j\alpha_i\alpha_jK(X_i,X_j)   \\
{\rm subject}\; {\rm to} & \sum_{i}y_i \alpha_i = 0, \\
& 0 \leq \alpha_i \leq C   \;\;\; \mbox{for}~ i=1,
\ldots,n.
\end{array}
\end{equation}
where $K(\cdot,\cdot)$ is the kernel function associated to 
the transformation $\phi$, that is,
$K(x,z)=\left\langle\phi(x),\phi(z)\right\rangle$.
In the dual problem for the soft margin $L^2$ problem, the last 
restriction in (\ref{sml1dual}),  $\alpha_i \leq C$, does not appear. This fact simplifies the
 theoretical analysis presented below in the $L^2$ case.
The classifier corresponding to the solution of (\ref{sml1dual}) is 
\beq\label{class}
\mbox{class}(x)=\mbox{sgn}\left(\sum_{i}y_i\alpha_i^*K(X_i,x)+b^*\right).
\enq 

The estimated prediction error for a given data set is the percentage of points from a test set that are incorrectly predicted. 
Let us notice that $\alpha_i^*$ values greater than zero
are the only ones that matter for the classifier. The corresponding feature vectors $X_i$ are the so-called 
{\it support vectors}. Low cost ways of finding or approximating these vectors
is a key issue addressed in the present article.

From the computational viewpoint, solving the convex quadratic program given by (\ref{sml1dual}) 
can be demanding when the data set is large. A good part
of the SVM literature is devoted to finding efficient ways to solve this
problem. Some of the ideas that have emerged as a result have to do
with solving appropriate sub-problems. In this direction, Osuna et al.
\cite{Osuna} introduce a decomposition algorithm in which the original 
quadratic programming problem is replaced by a sequence of problems of smaller 
size. An initial working set is considered and data are discarded from the
working set or included  from the remaining data depending on 
conditions of optimality. This method relates to the ``chunking'' methodology
as explained in \cite{Vapnik98} and to the SVM$^{light}$ algorithm of Joachims 
\cite{Joachims}. 

The Sequential Minimal Optimization (SMO) method of Platt \cite{Platt} is a fairly successful proposal along the line of working with smaller subproblems. In
SMO the optimization problem in (\ref{sml1dual}) is solved for two of the
$\alpha_i$ multipliers at a time, a simplification that allows a closed
form solution for each step.  
In a different direction, Mangasarian and Musicant \cite{Manga} consider the 
application of \emph{Succesive Overrelaxation} (SOR), a method developed 
originally to solve symmetrical linear complementarity and 
quadratic programs in order to obtain the 
solution of the SVM problem. 

More in the spirit of identifying the support vectors of the
training sample, therefore reducing the training set size, the proposal by Almeida et al\cite{Barros} applies the $k$-means 
clustering algorithm to the training set and evaluates the resulting
clusters for homogeneity. If a cluster is relatively pure (most observations 
coming from a single class), its data are discarded. But if a cluster contains 
data from different classes, it is assumed that,  there are probably support 
vectors in this cluster. Then the 
observations in the cluster are used in the reduced training data.
Abe and Inoue \cite{Abe} propose computing a Mahalanobis distance
classifier first and use those data points, misclassified by this preliminary
procedure, in the reduced training set. This assumes that those misclassified
by the Mahalanobis procedure are more likely to be near the boundary separating 
surface for the classification problem. In \cite{Shin03}, \cite{ShinHMN} and 
\cite{Shin07} the idea of identifying probable support vectors is
developed around neighbor properties. Those feature vectors
whose neighbors are mixed,  belonging to different classes,
as well as their nearest neighbors, are included in the reduced training set.


In this order of ideas, Camelo, Gonz\'alez-Lima and Quiroz \cite{Camelo} presented a 
subsampling and nearest neighbors method in which the support vector
set of the solution to the SVM problem for a small subsample
is iteratively enriched with
nearest neighbors to build a relevant training set significantly smaller
than the original training data set. The present work seeks to improve the results in \cite{Camelo} 
by the use of bagging and importance sampling. The idea is to enrich the subsample with more candidates to support vectors by looking simultaneously in different samples and searching for neighbors according to the intensity of candidates found. Since our proposal is closely related to the one in \cite{Camelo}, their method will be described in more detail in the following section.

The rest of this paper is organized as follows. In Section 2, we motivate and 
describe our approach  based on bagging and importance sampling. In Section 3, we report and discuss the results of applying the proposed methodology on benchmark examples. Section 5 contains
the proof of theoretical results that support the use of
subsampling methods, complementing those presented in \cite{Camelo}. 
The last section includes some conclusions and remarks.

\section{Bagging and importance sampling for support vectors}

The proposal in \cite{Camelo}, based on subsamples and nearest neighbors, can be summarized in the following steps:

{\it Procedure CGLQ}
\begin{itemize}
	\item[(i)] Select a random sub-sample consisting of a small fraction of the 
	set of examples. 
	\item[(ii)] Solve the SVM problem in that sub-sample, that is, 
	identify the support vectors for the sub-sample. Denote as
	$\cal V$ this initial set of support vectors. 
	Evaluate the error of the classifier on a test sample set.
	\item[(iii)] $\cal V$ is enriched with the  $k$-nearest neighbors (of each 
	element of $\cal V$) in the complete sample.
	$\cal V$ is also enriched with the addition of a new small random subsample 
	of the complete sample. With this \textit{new sub-sample}, we return to 
	step (ii). The iteration stops when there is no important improvement in the classification error.
\end{itemize} 
As reported in \cite{Camelo}, on quite diverse benchmark examples, this 
procedure achieves a classification error comparable to that corresponding to 
solving the SVM problem on the (original) complete training sample, with significant savings on computation time. 

One interesting feature of the method proposed in \cite{Camelo} is that
the methodology is not tied to a particular way of solving the SVM
problem on the training data. 

On the original idea of \cite{Camelo} one could 
consider the following modifications:
\begin{itemize}
	\item[1.] The initial sub-sample, in step (i) of procedure CGLQ
	can be substituted by a number of small sub-samples, solving 
	the SVM problem on each one. In this way, we
	would have a richer supply of candidates to approximate support vectors. The
	idea of applying an statistical learning 
	procedure to ``bootstrap samples'' taken
	from the original training sample, and then combine the output of
	those different adjusted predictors, is called {\it bagging} and was originally
	proposed by Breiman (see \cite{Breiman} and \cite{Breiman2}). Bagging has
	been particularly successful in the classification context, specifically
	when used on CART. In that case, the idea of bagging is called 
	Random Forests, and quite often improves on the performance of 
	a single classification tree. Certainly, a trade off must be established
	in the present context for the application of bagging. If too many initial
	samples are considered, a rich set of approximate support vectors will
	be available, but the savings in computation time could be lost.
	
	\item[2.] The enriching by nearest neighbors can be improved if a certain
	``intensity of support vectors by region'' can be estimated at each point
	of interest and a sampling procedure is used that takes this intensity into account in order to 
	sample more heavily in regions where more support vectors should be expected. In Monte Carlo 
	simulation, this type of idea 
	appear in what is called {\it importance sampling}. See \cite{Dunn}, for
	instance.
	
\end{itemize}
It turns out that, by the use of bagging, the goal of estimating a
local intensity of support vectors can be achieved.  Our proposal in
this direction will be to sample and add (to the original set
of support vectors) more sample points in those regions where more support vectors are expected.
These ideas are embodied in the following procedure. 
We slightly part from the usual 
bootstrap practice, by making our initial small sub-samples disjoint (we are
not sampling with replacement as in the original definition of bootstrap). 
This is done to produce a larger initial set of near support vectors. 
As in the introduction, the training sample is of size $n$ and
is formed by pairs $(X_i,y_i)$ of feature vectors and class variables,
with $y_i \in\{-1,1\}$. 
{\it Procedure Local sampling SVM}
\begin{itemize}
	\item[(i)] For a positive integer $L$ and $0<\delta<1$, from the original 
	sample, ${\cal X}$,   
	select $L$ disjoint subsamples  ${\cal T}_{i}$, $i=1,\ldots,L$, 
	each of size $n_S$, such 
	that $n_S=\lfloor \delta n / L\rfloor$, where $\lfloor\cdot\rfloor$
	denotes the {\it floor} function.
	We denote as $\mathcal{T}$ the set 
	of observations of all the sub-samples i.e. $\mathcal{T}=\cup_{i=1}^{L}
	{\cal T}_{i}$. $\mathcal{T}$ represents a fraction $\delta$ of the entire 
	training sample.     
	\item[(ii)] On each subsample ${\cal T}_{i}$, solve the SVM problem, 
	finding the set $\mathcal{V}_{i}$, of support vectors associated to that
	sub-sample. Let ${\cal V}$ denote the union of these initial support 
	vector sets, i.e. 
	${\cal V}=\cup_{i=1}^L{\mathcal{V}}_{i}$ and let $m$ be the size 
	(cardinality) of ${\cal V}$.
	\item[(iii)] Let $k=\lfloor\ln m\rfloor$.
	For each 
	$\nu_{j} \in {\cal V}$, identify its $k$th-
	nearest-neighbor in ${\cal V}$. Denote this $k$th-nearest-neighbor
	by $NN_k(\nu_j,{\cal V})$ and denote by $\rho_j$ the 
	distance between $\nu_{j}$ and $NN_k(\nu_j,{\cal V})$: $\rho_j=
	\dis(\nu_j,NN_k(\nu_j,{\cal V}))$, where 
	$\dis(\cdot,\cdot)$ stands for Euclidean distance. Let 
	$\rho$ denote the median of the radii $\rho_j$.
	\item[(iv)] For a parameter $\beta>0$, let $r=\beta\rho$.
	Define
	\[
	\eta_j=\frac{\rho_j^{-1}}{\sum_{i=1}^{\#{\cal V}}\rho_i^{-1}}
	\] 
	Sample a fraction $\eta_j$ of the points of 
	${\cal X}\setminus\cal T$ in the ball with center $\nu_j$ and 
	radius $r$. Write $\cald_j$ for this random sample. 
	\item[(v)] Solve the SVM problem for the new sample 
	\[
	{\cal V}\cup \left(\cup_{j=1}^L{\cald}_{j}\right).
	\]
	
\end{itemize}

Three observations are necessary regarding the procedure just described:

\noindent 1. In order to explain the way importance sample is approximately
implemented in our procedure, let us recall the idea of density
estimation by ``Parzen windows'' (see Section 4.3 in \cite{Duda} for details, including
a proof of consistency). Given an i.i.d. sample, $X_i,\dots,X_n$ in $\real^d$, obtained
from a probability distribution that admits a density $f(\cdot)$, 
for an arbitrary $x\in \real^d$ (which could be one of the sample points), if
$r_k(x)$ denotes the Euclidean distance from $x$ to its $k$-th nearest neighbor in the sample,
then the density $f(x)$ is consistently estimated by a constant times the reciprocal of
the volume of the $k$-th nearest neighbor ball. This means that $f(x)$ is proportional
to $(r_k(x))^{-d}$. In our case, we use the distance $\rho_j$ between each $\nu_{j}$ and
its $k$-th nearest neighbor in ${\cal V}$ to get an estimate of the ``support vector
density'' near $\nu_{j}$. Then we set a fixed radius $r$ and sample in a ball of
radius $r$ around $\nu_{j}$ with an intensity proportional
to $\rho^{-1}_j$.
Precise importance sampling would require to sample with an intensity proportional
to $\rho^{-d}_j$, but preliminary experiments revealed
that this ``exact'' importance sampling would result too extreme in the sense of producing
heavy sampling in some regions and almost no sampling in others. For this reason, our
sampling is proportional to $\rho_j^{-1}$.

\noindent 2. An important difference with the approach put forward in \cite{Camelo} is that in that
approach, the enriching of the set ${\cal V}$ occurs inside the $k$-th nearest neighbor
balls, while in the present proposal, we use a common fixed radius and change the sampling
intensity at each $\nu_{j}$.

\noindent 3. The parameter $\beta$ in the procedure just described 
provides flexibility by allowing the user to
vary the radius (and volume) of the balls in which the sampling is performed.

Next, we study the behavior of the Local Sampling procedure applied to some real life problems.
	
	\section{Applications in benchmark data}

In this section we present a performance evaluation of the methodology proposed on four datasets from the LibSVM library\footnote{See \texttt{https://www.csie.ntu.edu.tw/~cjlin/libsvm}}. For the
experiments described in this section we have used the statistical software \texttt{R} and the \texttt{e1071} package\footnote{For more details visit \texttt{https://cran.r-project.org/web/packages/e1071/}}. Procedures are run on a computer with Motherboard EVGA Classified SR-2 with two microprocessors @ 2.67 GHz and RAM memory of 48 GB  1333 MHz. 

For our experiments we consider the following three kernels for the SVMs:

\begin{itemize}
	\item Linear: $K(\ve{x},\ve{x}_{j})=\ve{x}_{j}^{T}\ve{x}$.
	\item Polynomial with degree $p$, $p\in\mathbb{N}$: $K(\ve{x},\ve{x}_{j}$)=$(\gamma \ve{x}_j^{T}\ve{x})^p$.
	\item Radial basis: $K(\ve{x},\ve{x}_{j})=\exp(-\gamma \|\ve{x}-\ve{x}_{j}\|^{2})$.
\end{itemize}

Table 1 displays the description of the datasets tested.  These examples cover different ranges
with respect to sample size and data dimension. 

\vspace{15 pt}

\begin{table}[h]
	\label{table:dataset}
	\centering
	\begin{tabular}{|c c c c|}
		\hline\hline		
		Dataset & TrainSize &  TestSize & Features    \\ [0.5ex]
		\hline
		COD-RNA	& 59,535  & 20,000  & 8\\
		IJCNN1 & 49,990 & 15,000 & 22 \\
		COVTYPE	& 521,012 & 60,000 & 54  \\
		WEBSPAM &  300,000 & 50,000 & 254 \\
		\hline
	\end{tabular}
	\caption{ Datasets description}
\end{table}

 Section 3.2 encompasses the results obtained for the problems tested using the three different kernels. For each one of them, a description is included on how the Local Sampling Algorithm is applied and three tables are displayed. One of them shows the results obtained when solving the problem using the complete data set. Information on the classification error, number of support vectors and computational time are included. The other two tables show the results obtained when using the Local Sampling approach and the CGLQ algorithm from \cite{Camelo}. Each of these tables contain the percentage of the support vectors found by the subsample procedures that are support vectors for the full dataset and  the ratio between the classification errors corresponding to the subsample procedures and the full data set. A value less than one for this quotient indicates an improvement in the error when using the subsample approach. The tables also include the percentage that the computational time for the subsample procedures represents of the running time for the complete data set.     

\subsection{Parameter Choices }

In order to choose the kernel parameters and the cost parameter $C$ for the optimization problem, 
10-Fold Cross-Validation was used on a small random sample corresponding approximately to 1\% of 
the full training data. The parameters obtained for each problem and kernel are the ones used in 
all our tests.

With the parameters $\gamma$ and $C$ chosen, the SVM classifier was fitted on the full training sample, 
obtaining the number and identity of support vectors, the execution time and the test error.

For the Local Sampling approach we proceeded as follows. For each combination of data set and kernel, a fraction $\delta$ (this $\delta$ depends on the
problem) of the training data was selected and divided into $L$  sub-samples. The Local Sampling SVM procedure from the previous section was run initially for a value of
$\beta$ equal to 0.1 and repeated, increasing $\beta$ by 0.1 at each iteration until there is no 
decrease in the classification error. 
Results for the final value of $\beta$ are reported, except that
the evaluation done to search for the optimal value of $\beta$ is included in the execution time reported.
We take note of the initial amount of support vectors, which is the cardinality of ${\cal V}$ at the 
end of step (ii) of the procedure, the number of support vectors at the end of the procedure, the 
number of these which are support vectors for the problem solved on the full training data set, the 
generalization 
error on the test data and the execution times. 

This whole procedure is run 10 times, on independently chosen initial samples, in order to report averages of the amounts of interest over independent realizations. The next subsection includes the results obtained for the data sets tested. 
\subsection{ Problems Tested}
\subsubsection*{COD-RNA Dataset}

The COD-RNA data set contains data for the detection of non-coding RNA's on the basis of predicted 
secondary structure. 
The data are divided into 59,535 observations in the training set and we consider 20,000 observations 
in the test sample, each with 8 attributes measuring genetic information.

Table \ref{tab:fullcodrna} displays the results for the SVM solution considering the full training sample and the optimal parameters $C$ and $\gamma$ obtained in the preliminary evaluation for each of the kernels considered. 


\begin{table}[h!]
	\caption{SVM results from full COD-RNA training data}
	\small
	\label{tab:fullcodrna}
	\centering
	\begin{tabular}{|c |c c| c c c|}
		\hline\hline		
		& \multicolumn{2}{c|}{Optimal parameters} & &   &     \\ [0.5ex]
		Kernel  &  Cost &  $\gamma$  & Number of Support Vectors  & Error rate & Time  (s) \\
		\hline
		Linear	&    0.1 & -- & 22923 (11462 $||$ 11461) & \textbf{0.206}   & 178.013   \\
		Polynomial	& 	 1 &  0.1 & 15397 (7698 $||$ 7699) &  0.208 & \textbf{114.115} \\
		Radial basis  &  10 & 0.1  &  8768 (4380 $||$ 4388) & 0.242 & 117.901  \\
		\hline
	\end{tabular}
\end{table}

The second and third columns in Table \ref{tab:fullcodrna} show the optimal parameters obtained by cross validation. The following column shows the number of support vectors found in
the solution for each kernel. The quantities in parentheses that follow, indicate the division of 
these support vectors by class. The error rates obtained in the test set are included in the fifth 
column. For the polynomial kernel, in this example, the degree used was $p=3$.

In this case, the best performance for the complete data set, in terms of test error, is obtained 
for the linear and polynomial kernels, which produce very similar errors. The linear kernel solution 
requires a significantly larger amount (22,923) of support vectors, compared to the other 
kernels. Execution times in seconds are shown in the sixth column. The polynomial kernel has the best 
performance in terms of execution time, taking approximately 114 seconds. The higher execution
time needed by the linear kernel solution probably correlates with the large number of support 
vectors this solution requires.

To apply the local sampling SVM scheme to the CODRNA data, we used $\delta=0.01$, dividing the 
training sample in 12 subsamples of size 50.  With the optimal parameters chosen and for each
of the three kernels, we executed 10 runs of the algorithm proposed as described above.
Table \ref{tab:codcam} shows the results for our methodology compared with the results obtained using the methodology proposed in \cite{Camelo}. As pointed out before the
results are shown only for the final value of $\beta$. 

\vspace{15 pt}

\begin{table}[htp!]
	\centering
	\caption{COD-RNA error and computing times vs CGLQ algorithm}
	\label{tab:codcam}
	\begin{subtable}{0.5\linewidth}
		\centering
		\scriptsize \begin{tabular}{|l|*{3}{c|}}
			\hline
			\textbf{COD-RNA} & \multicolumn{3}{c|}{Kernel} \\ 
			\cline{2-4}
			& Linear & Polynomial  & Radial basis \\
			\hline
			$\beta$  & 0.1 & 0.1 & 0.1 \\
			SV initial & 395 & 442 & 362  \\
			SV final & 1296 & 1293 & 585 \\
			SV real & 1127 & 757 & 338 \\
			\% full SV & 4.91 \% & 4.91\% & 3.85 \% \\
			Error rate & 0.21  & 0.19  & 0.24  \\
			Sd. dev. & (0.02) & (0.007) & (0.01) \\
			Error ratio & 1.02 &  0.92 &  1.01 \\
			Time (s) & 1.1095  & 1.3719 &  0.8647 \\
			\% full time   & 0.62  \% & 1.2 \% & 0.73 \%\\
			\hline
		\end{tabular} 
		\caption{\scriptsize Local Sampling SVM}
		\label{tab:cod1}
	\end{subtable}%
	\hfill
	\begin{subtable}{0.5\linewidth}
		\centering
		\scriptsize\begin{tabular}{|l|*{3}{c|}}
			\hline
			\textbf{COD-RNA} & \multicolumn{3}{c|}{Kernel} \\ 
			\cline{2-4}
			$\delta$=0.01, $K$=5	& Linear & Polynomial  & Radial basis \\
			\hline
			&  &  &  \\
			SV initial & 272 & 320 & 193  \\
			SV final & 3417 & 2945 & 1695 \\
			SV real & 3332 & 2203 & 1184 \\
			\% full SV & 14.53 \% & 14.3 \% & 13.51 \% \\
			Error rate & 0.2068  & 0.193 & 0.24  \\
			Sd. dev. & (0.005) & (0.005) & (0.01) \\
			Error ratio & 1.004 &  0.92 &  1.01 \\
			Time (s) & 5.08  & 4.89	 &  5.03 \\
			\%  full time   & 2.85  \% & 4.29 \% & 4.26 \%\\
			\hline
		\end{tabular} 
		\caption{\scriptsize CGLQ Algorithm}
	\end{subtable}%
\end{table}

In Table \ref{tab:codcam} we observe the following: The method proposed in \cite{Camelo} finds
a significantly larger number of complete sample support vectors, in comparison with the
local sample procedure proposed here. Still, in terms of test errors the results for the
method of \cite{Camelo} and our local sample scheme are very similar and both are quite successful,
with error ratios very close to 1 for all the kernels. The main difference between the two
methods compared lie, in this example, in the execution time. For all kernels, the method
proposed here requires between 20\% and 25\%  of the computing time in comparison to the algorithm
of \cite{Camelo} (that was already very efficient in terms of computing time).

\subsection*{IJCNN1 Dataset}

The IJCNN1 data set was used in the 2001 Neural Network Competition, and contains information on a time series produced by a physical system. It consists of 49,990 observations in the training sample and we use 15,000 observations for testing. The number of features is 22.

For this data set, the degree used for the polynomial kernel was $p=3$. The results obtained
for the SVM problem on the complete training data set using the estimated optimal
parameters are shown in Table \ref{table:fullijcnn}.

\vspace{10pt}

\begin{table}[h]
	\caption{SVM results from full IJCNN1 training data}
	\small
	\label{table:fullijcnn}
	\centering
	\begin{tabular}{|c |cc| c c c|}
		\hline\hline		
		& \multicolumn{2}{c|}{Optimal parameters} & &   &     \\ [0.5ex]
		Kernel  &  Cost &  $\gamma$  & Number of Support Vectors  & Error rate & Time  (s) \\
		\hline
		Linear	& 5 & -- & 8577 (4296 $||$ 4281) & 0.078   & 63.99  \\
		Polynomial	&  0.1 &  0.5 & 9377 (4706 $||$ 4631) & 0.071 & 62.5 \\
		Radial basis &  10 & 0.1 & 5731 (2877 $||$ 2854) & \textbf{0.028} & 63.45 \\
		\hline
	\end{tabular}
\end{table}

On this example, the execution times on the complete data set are very similar for the
three kernels. In terms of test error, the best performance is obtained with the
radial basis kernel, reaching an error rate of 2.8 \%. 

In this case, the local sampling SVM approach is used with 20 subsamples of size 50, 
corresponding to a value of $\delta$ equal to 0.02. Tables \ref{tab:ijcnn1} and \ref{tab:ijcnn2} summarize the
results on the IJCNN1 for the Local Sampling procedure and the CGLQ method.

\vspace{10 pt}

\begin{table}[htp!]
	\centering
	\begin{subtable}{0.5\linewidth}
		\centering
		\scriptsize \begin{tabular}{|l|*{3}{c|}}
			\hline
			\textbf{IJCNN1} & \multicolumn{3}{c|}{Kernel} \\ 
			\cline{2-4}
			& Linear & Polynomial  & Radial basis \\
			\hline
			$\beta$  & 0.2 & 0.2 & 0.2 \\
			SV initial & 347 & 503 & 395  \\
			SV final & 834 & 407 & 1055 \\
			SV real & 711 & 317 & 610 \\
			\% full SV & 8.29 \% & 3.38 \% & 10.64 \% \\
			Error rate & 0.069   & 0.06  & 0.028  \\
			Sd. dev. & (0.002) & (0.008) & (0.001) \\
			Error ratio & 0.89 &  0.85 &  1.02 \\
			Time (s) & 1.98  & 2.18 &  6.23 \\
			\%  time  SVM & 3.09  \% & 3.48 \% & 9.82 \%\\
			\hline
		\end{tabular} 
		\caption{\scriptsize Local Sampling SVM}
		\label{tab:ijcnn1}
	\end{subtable}%
	\hfill
	\begin{subtable}{0.5\linewidth}
		\centering
		\scriptsize\begin{tabular}{|l|*{3}{c|}}
			\hline
			\textbf{IJCNN1} & \multicolumn{3}{c|}{Kernel} \\ 
			\cline{2-4}
			$\delta$=0.02, $K=$5	& Linear & Polynomial  & Radial basis \\
			\hline
			& &  &  \\
			SV initial & 180 & 263 & 197  \\
			SV final & 673 & 295 & 394 \\
			SV real & 607 & 235 & 234 \\
			\% full SV & 7.07 \% & 2.5 \% & 4.09 \% \\
			Error rate & 0.068   & 0.07  & 0.036  \\
			Sd. dev. & (0.005) & (0.01) & (0.0008) \\
			Error ratio & 0.87 &  1.08 &  1.29 \\
			Time (s) & 2.5  & 0.77 &  2.98 \\
			\%  time  SVM & 3.68  \% & 1.23 \% & 4.78 \%\\
			\hline
		\end{tabular} 
		\caption{\scriptsize Camelo et al. Algorithm}
		\label{tab:ijcnn2}
	\end{subtable}%
	\caption{IJCNN1 error and computing times}
\end{table}

Table \ref{tab:ijcnn1} shows that, for the IJCNN1 data set, the local sampling scheme finds a larger
fraction of the full data support vectors, in comparison to the procedure of \cite{Camelo}. Both
methods perform very well and similarly in terms of test error for the Linear and Polynomial
kernels. For the Radial Basis kernel, which was the kernel with the smallest error on
the complete sample, the performance of local sampling is better, keeping
the error ratio near 1. This comes at the cost of doubling the execution time of 
procedure CGLQ, but
still keeping the execution time under 10\% of that for the full data set.


\subsection*{COVTYPE Dataset}

The Covertype data set contains  information for predicting forest cover type from seven cartographic variables. In this case, we considered a modified version that converts a seven class problem into a binary classification problem, where the goal is to separate class 2 from the other 6 classes. The number of instances is 581,012 and each instance is described by 54 input features. For our implementation we consider 521,012 observations in the training set and 60,000 instances for testing.

For this data set, the degree used for the polynomial kernel was $p=2$. Table \ref{tab:fullcovtype} shows the results of solving this problem on the complete training
data set, for the optimal parameters chosen. For all three kernels, there is a
large amount of support vectors, evenly distributed between the two classes. In terms of error, 
the best performance is obtained with the 
radial basis kernel with an error rate equal to 9 \%.  This is a complex problem, for which
the best solution takes about eight hours of computation time. 

\begin{table}[h]
	\caption{SVM results from full COVTYPE training data}
	\small
	\label{tab:fullcovtype}
	\centering
	\begin{tabular}{|c cc| c c c|}
		\hline\hline		
		& \multicolumn{2}{c|}{Optimal parameters} & &   &     \\ [0.5ex]
		Kernel  &  Cost &  $\gamma$  & Number of Support Vectors  & Error rate & Time  (s) \\
		\hline
		Linear	&   10 & --  & 303036 (151517 $||$ 151519) & 0.237   & 36688.26  \\
		Polynomial	 	& 1 &  1 & 216841 (108421 $||$ 108420) & 0.169 & 39896.5 \\
		Radial basis &  10 & 5 &  144271 (72113 $||$ 72158) & 0.09& 28746.54 \\
		\hline
	\end{tabular}
	
\end{table}

For the local sampling SVM approach, we used 50 subsamples of size 250, corresponding to
a choice of $\delta=$0.025. Tables \ref{tab:cov1} and \ref{tab:cov2} show the comparison of
results in this example for the 
local sampling and the CGLQ method. 

In this case, the optimal value of the parameter $\beta$ goes up to 0.5 in order to reach
better error rates. This shows the usefulness of a flexible parameter $\beta$ in complex problems.
The local sampling procedure identifies as support vectors a large fraction (about 40\%) of
the support vectors for the complete problem.
The error rates achieved by both methods compared are very good and quite similar. The main
difference in performance is again in the execution times, which are significantly 
smaller for the local sampling procedure.

\begin{table}[htp!]
	\centering
	\begin{subtable}{0.5\linewidth}
		\centering
		\scriptsize 
		\begin{tabular}{|l|*{3}{c|}}
			\hline
			\textbf{COVTYPE} & \multicolumn{3}{c|}{Kernel} \\ 
			\cline{2-4}
			& Linear & Polynomial  & Radial basis \\
			\hline
			$\beta$  & 0.5 & 0.5 & 0.5 \\
			SV initial & 7779 & 7208 & 10785  \\
			SV final & 139747 & 96003 & 72188 \\
			SV real & 136108 & 87888 & 63103 \\
			\% full SV & 44.91 \% & 40.53\% & 43.73 \% \\
			Error rate & 0.2391   & 0.1789  & 0.11  \\
			Sd. dev. & (0.0008) & (0.0006) & (0.001) \\
			Error ratio & 1.009 &  1.05 &  1.22 \\
			Time (s) & 6531.96 & 7928.2 &  7932.9 \\
			\%  time  SVM & 17.8  \% & 19.87 \% & 27.59 \%\\
			\hline
		\end{tabular} 
		\caption{\scriptsize Local Sampling SVM}
		\label{tab:cov1}
	\end{subtable}%
	\hfill
	\begin{subtable}{0.5\linewidth}
		\centering
		\scriptsize
		\begin{tabular}{|l|*{3}{c|}}
			\hline
			\textbf{COVTYPE} & \multicolumn{3}{c|}{Kernel} \\ 
			\cline{2-4}
			$\delta$=0.05, $K=$5 & Linear & Polynomial  & Radial basis \\
			\hline
			& &  &  \\
			SV initial & 15226 & 11974 & 10603  \\
			SV final & 76972 & 55204 & 37049 \\
			SV real & 76307 & 40306 & 21153 \\
			\% full SV & 25.18 \% & 18.58\% & 14.66 \% \\
			Error rate & 0.23   & 0.17 & 0.11  \\
			Sd. dev. & (0.00002) & (0.0009) & (0.01) \\
			Error ratio & 0.99 & 1.02 &  1.22 \\
			Time (s) & 11424.73 & 13343.96 & 8887.25 \\
			\%  time  SVM & 31.14  \% & 33.44 \% & 30.91 \%\\
			\hline
		\end{tabular} 
		\caption{\scriptsize Camelo et al. Algorithm}
		\label{tab:cov2}
	\end{subtable}%
	\caption{COVTYPE error and computing times}
\end{table}


\subsection*{WEBSPAM Dataset}

The Webspam data set is a collection of 350,000 webspam pages that were obtained using an automatic methodology. We consider the set used in the Pascal Large Scale Learning Challenge with the normalized unigram data set that contains 254 features. In our implementation, we split the data into two sets: training set with 300,000 observations and testing set with 50,000 observations.

The parameters $\gamma$ and $C$ were found as in the previous problems by using cross validation. The polynomial kernel was considered with degree $p=3$. Table \ref{tab:fullwebspam} shows the results for the solution of this problem on the
complete training data.  It appears that this is a problem in which the SVM methodology works well, in 
the sense that very low error rates are obtained, being the minimum test error of 1.1\%, obtained 
with the radial basis kernel. The linear kernel is probably the least convenient for this
problem, since it produces the largest error rate with the highest computing time. 


\begin{table}[h]
	\caption{SVM results from full WEBSPAM training data}
	\small
	\label{tab:fullwebspam}
	\centering
	\begin{tabular}{|c |cc| c c c|}
		\hline\hline		
		& \multicolumn{2}{c|}{Optimal parameters} & &   &     \\ [0.5ex]
		Kernel  &  Cost &  $\gamma$  & Number of Support Vectors  & Error rate & Time  (s) \\
		\hline
		Linear	&  10 & -- & 56810 (28407 $||$ 28403) & {0.068}   & 21917.83  \\
		Polynomial	 	& 10 &  2  & 16789 (8541 $||$ 8248) &  0.014 & 16492.46 \\
		Radial &   10 & 2 & 16623 (8748 $||$ 7875) & 0.011 & 10972.71  \\
		\hline
	\end{tabular}
\end{table}

For this data set, when the Local Sampling approach was used, it was necessary to choose a value of $\delta=0.1$ in order to obtain an approximation close enough to the best solution. For this value of $\delta$,  25 subsamples of size 1400 were used. Table~\ref{tab:web1} summarizes the results on this data set for the Local Sampling procedure.

\begin{table}[htp!]
	\centering
	\begin{subtable}{0.5\linewidth}
		\centering
		\scriptsize\begin{tabular}{|l|*{3}{c|}}
			\hline
			\textbf{WEBSPAM} & \multicolumn{3}{c|}{Kernel} \\ 
			\cline{2-4}
			& Linear & Polynomial  & Radial basis \\
			\hline
			$\beta$  & 0.5 & 1 & 1 \\
			SV initial & 9341 & 6083 & 7499  \\
			SV final & 35155 & 10141 & 13030 \\
			SV real & 32182 & 8755 & 9049 \\
			\% full SV & 56.46 \% & 52.04 \% & 54.46 \% \\
			Error rate & 0.069   & 0.0178  & 0.018  \\
			Sd. dev. & (0.0004) & (0.0005) & (0.0004) \\
			Error ratio & 1.02 &  1.27 &  1.63 \\
			Time (s) & 3931.96  & 3491 &  3683.11 \\
			\%  time  SVM & 17.93  \% & 21.16 \% & 33.56 \%\\
			\hline
		\end{tabular} 
		\caption{\scriptsize Local Sampling SVM}
		\label{tab:web1}
	\end{subtable}%
	\hfill
	\begin{subtable}{0.5\linewidth}
		\centering
		\scriptsize
		\begin{tabular}{|l|*{3}{c|}}
			\hline
			\textbf{WEBSPAM} & \multicolumn{3}{c|}{Kernel} \\ 
			\cline{2-4}
			$\delta$=0.10, $K$=10 & Linear & Polynomial  & Radial basis \\
			\hline
			&  &  &  \\
			SV initial & 6358 & 2770 & 3474  \\
			SV final & 31791 & 7764 & 10337 \\
			SV real & 29793 & 6208 & 6444 \\
			\% full SV & 52.44 \% & 36.97\% & 38.76 \% \\
			Error rate & 0.068  & 0.0169  & 0.017  \\
			Sd. dev. & (0.0004) & (0.0004) & (0.0002) \\
			Error ratio & 1.005 &  1.21 &  1.55 \\
			Time (s) & 7318.467  & 3396.78 &  2314.489 \\
			\%  time  SVM & 33.39  \% & 20.59 \% & 21.09 \%\\
			\hline
		\end{tabular} 
		\caption{\scriptsize Camelo et al. Algorithm}
		\label{tab:web2}
	\end{subtable}%
	\caption{WEBSPAM error and computing times}
\end{table}
 
Figure 1 shows the evolution of the estimated error for the Local Sampling solution of the WEBSPAM problem, as a function of $\delta$. The dotted green and blue lines show, respectively, the largest and smallest errors achieved with the Local Sampling methodology, while the dotted red line corresponds to the estimated error for the solution for the whole data set. 

\begin{figure}[h!]
	\centering
	\includegraphics[scale=0.65]{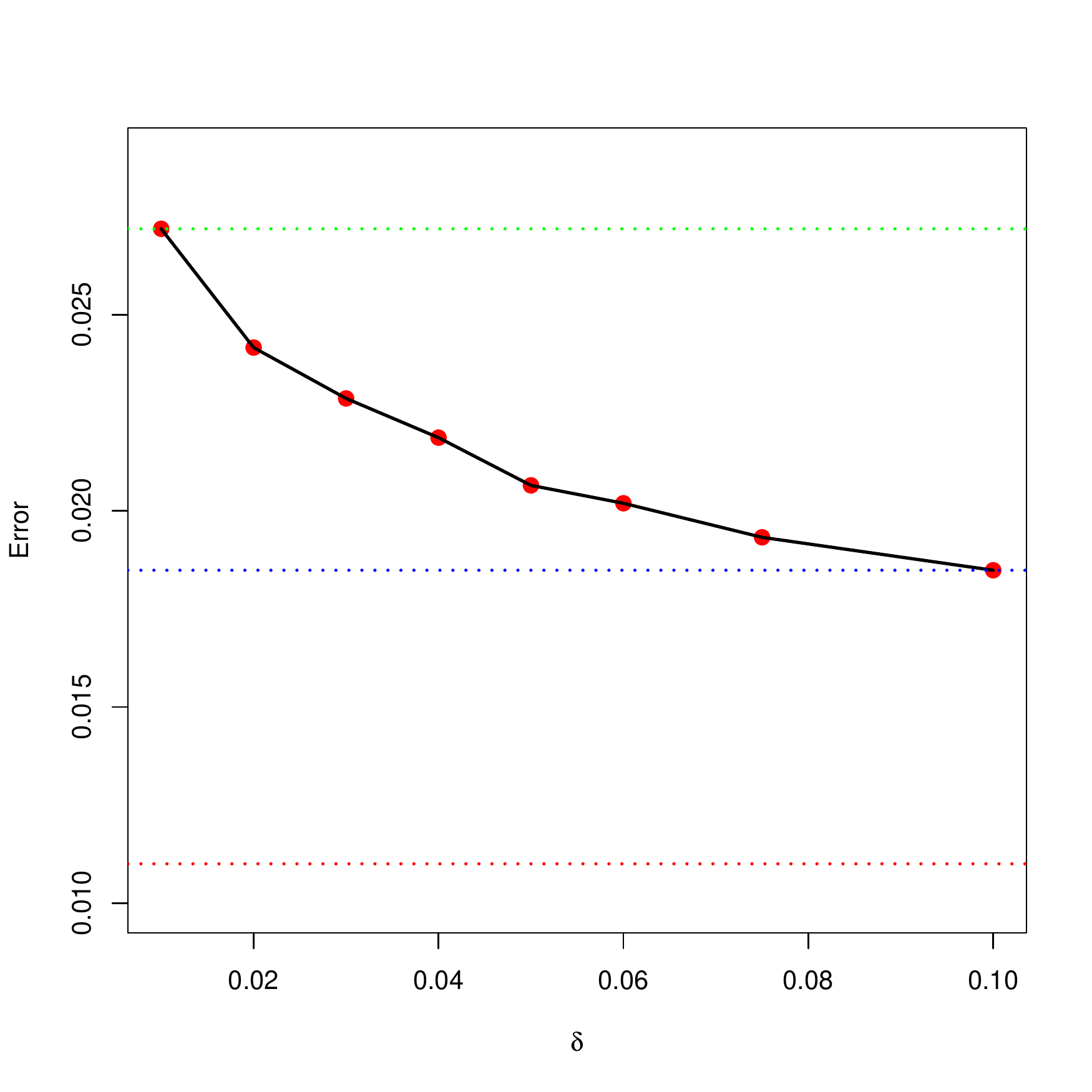}
	\vspace{5pt}
	\caption{Classification error vs $\delta \in [0.01,0.1]$ for the Local Sampling SVM procedure- WEBSPAM problem.}
	\label{fig:delta}
\end{figure}

Table~\ref{tab:web2} summarizes the results obtained with the CGLQ methodology. Comparing both subsample approaches for this example we find that, for the linear Kernel, the Local Sampling approach is better than CGLQ since it obtains similar error with a much lower computational time. And this error is almost the same as the one obtained for the whole data set. For the polynomial kernel, similar performances were found. However, for the radial basis kernel, the best results are obtained with the CGLQ algorithm,  in terms of the generalization error and the execution time. However, it should be highlighted that in this problem, for the nonlinear kernels, the classification errors reached by any of the subsample procedures are, in terms of the error ratio, the worst  among all the tested examples. We believe that this performance is due to the detrimental effect that may have the high dimensions of the data when using these type of approaches, as our theoretical results suggest. This will be discussed in Section 4.

\subsection{Practical Remarks}

In this subsection we would like to summarize our findings regarding the parameter choices ($\delta$, $L$, and $\beta$) for the Local Sampling procedure, which could serve as guidelines for its use. 

The size of each subsample is determined by the value of the parameter $\delta$. We conclude, from our experimentation, that for problems with data set of medium size, this is, size of order around $10^5$ and $10^6$, and low dimension (dimension less than 30), the values of delta should be chosen between $1\%$ and $5\%$. This can be implemented by starting from $\delta=1\%$ and sequentially increasing the value, depending on the improvement in the error and the execution times. For problems with larger training sample size in higher dimension, consider using larger values of delta, up to $10\%$. 

The number of subsamples, $L$, to be chosen depends on the value of $\delta$ and the complete data set size. If the number of points at each subsample is too small, this is, that there is poor information about the full population inside the subsamples, a large number of these points will become support vectors of the corresponding SVM problems and they will not contribute to approximate the real ones for the complete data set. Therefore, $L$ is chosen trying to get a balance between the number of observations defined by the $\delta$'s and the distribution of these observations in the subsamples, in such a way that the number of points at each subsample is not too small. We want to get a reasonable amount of support vectors for each subproblem that provide valuable information for the next step.

Regarding the choice of the parameter $\beta$, our experience indicates that values in the 
interval $[0.1, 0.5]$ are useful, in terms of including enough information from the sample. Increasing beta will improve the performance in terms of validation error at the cost of larger execution time. A guideline could be 
to start with $\beta=0.1$ and repeat with increments of $0.1$ while there is a significant improvement in classification error or until a limit is reached in terms of execution time.

Next we will present our theoretical results. 	
	
	\section{Theoretical results}
	
	The purpose of the present section is to establish a stability result for the solution of
	the soft margin $L^1$ SVM problem, that can shed some light on the performance of subsampling 
	schemes as the one proposed in the preceding sections.
	
	Using the notation established in Section 1, here we consider an i.i.d. sample of pairs
	$(X_i,y_i)$ for $i=1,\dots n$, where the $X_i$ follow
	a probability distribution $\mu$ on $\real^d$, the $y_i$ take the values
	$1$ and $-1$ and, given $X_i=x$, the distribution of $y_i$ is given by
	$\Pr(y_i=1\dado x)=\eta(x)$ for a measurable function $\eta$ on on the support of $\mu$.
	
	We assume that $\mu$ admits a density $f$ with respect to Lebesgue measure, $\lambda$,
	which is bounded away from zero and infinity on its compact support, $\cal S$. Also,
	we assume that $\eta$ (at least in the vicinity of support vectors), is bounded away from
	0 and 1. 
	
	A random subsample, $\cal M$, is taken, of size 
	$n'=\lceil \delta n \rceil$ for some $\delta>0$. Based on this subsample, the SVM problem is
	solved and  our purpose is to quantify the similarity of the solution obtained with
	$\calm$ to the one given by (\ref{class}), in which the whole data set is employed. In what 
	follows, the expression \emph{with high probability}, means that the probability of the 
	event considered goes to 1, as $n \to \infty$.
	
	Let ${\cal V}_{n}$ the set of support vectors for the SVM solution computed with the complete 
	sample of size $n$ and let $q$ be its cardinal, i.e.  $q=\#{\cal V}_{n}$, the number of 
	support vectors. The first thing to be verified is that with \emph{high probability}, the 
	subsample $\cal M$ will contain points close enough, say, to a distance less than $\rho$, of 
	each point $X_{i}$ in the distinguished set ${\cal V}_{n}$, for any $\rho$ such that 
	$\rho^{d}$ is of order $\mathrm{O}(\logn^2 n /n)$.
	
	Besides the assumptions on the sample distribution stated above, we will assume the following
	shape condition on 	$\cal S$, called \emph{weak grid compatibility},  which is a relaxed 
	version of a condition considered in \cite{Yukich}, in the context of theory of clustering 
	algorithms. 
	
	\begin{definition}
		\emph{Let $\cal S$, as above, denote the support of the density $f$ generating the data.
			We will say that $f$ and $\cal S$ satisfy the \emph{weak grid compatibility} condition, if there is a positive number $\gamma$, such that  for all small enough $l>0$, $\cal S$ can be covered, possibly after translation and rotation, by a regular array $\cal W$, of cubes of side $l$, such that:}	
		
		\begin{displaymath}
		\forall V \in {\cal W}, \lambda(V \cap {\cal S}) \geq \gamma \lambda(V),
		\end{displaymath}  	
	\end{definition}
	
	\noindent where $\lambda$ denotes Lebesgue measure in $\rea^{d}$ and  \emph{regular array 
	of cubes in $\rea^{d}$} means that the vertices of the cubes form a  regular grid in 
	$\rea^{d}$, in the sense that the length between the contiguous vertices is constant and equal 
	to $l$. 
	In this definition  it is assumed that $\cal W$ includes only the cubes needed to cover 
	$\cal S$, that is, if $W \cap {\cal S}= \emptyset$ the $ W \notin \cal W$. According to 
	\cite{Yukich}, $d$-dimensional balls and ellipsoids in $\rea^{d}$ as well as regular polyhedra 
	satisfy this condition.
	
	As a condition on the kernel $K(\cdot,\cdot)$ employed in the SVM procedure, we assume
	that $K$ is Lipschitz continuous, with Lipschitz constant $L$.
		
	Finally, we make assumptions on the
	solution of the soft margin problem (\ref{sml1prim}) on the whole data set.
	First we assume the following asymptotic
	continuity condition on the uncertainty of the classification function for the 
	solution of the SVM problem on the complete data set: Let  $\alpha_i^*$ and $b^*$
	be the solution coefficients appearing in (\ref{class}). We assume that
		
	\beq\label{asympcont}
	\lim_{\epsilon\to 0}\,\limsup_n\mu\left(x\in{\cal S}:\left|\sum_{i}y_i\alpha_i^*K(X_i,x)+b^*
	\right|
	\leq \epsilon\right)=0, a.s.
	\enq
	where the expression ``almost surely'' refers to the product space of infinite samples. 
	Condition (\ref{asympcont}) is bounding the level of uncertainty that the solution
	to the SVM problem admits. Points $x\in \cal S$ for which the condition 
	\[
	\left|\sum_{i}y_i\alpha_i^*K(X_i,x)+b^*\right|
	\leq \epsilon
	\]
	holds, are points near the boundary of indecision of the SVM solution. We are asking that, as 
	$\epsilon$ becomes smaller, this region of $\epsilon$-uncertainty has probability that goes
	to zero. Simulations shown in Subsection \ref{indecision} suggests that condition 
	(\ref{asympcont}) holds comfortably in real examples. 
	
	The last assumptions, on the other hand, are technical and more
	restrictive assumptions on the number of support 
	vectors that the solution associated to the complete data set
	might have. On one hand, we suppose that the cardinality $q$ of the set 
	${\cal V}_{n}$ of support vectors 
	is $\och\Big(\Big(\frac{n}{\logn^2 n}\Big)^{1/d}\Big)$, namely, for each
	$\epsilon>0$, 
	\beq\label{cond_card_1}
	\Pr\left(q >\epsilon\left(\frac{n}{\logn^2 n}\right)^{1/d}\right)\rightarrow 0, 
	\>\>\mbox{ as }\>\>n\rightarrow \infty.	
	\enq
	The power $1/d$ in this bound, makes the condition more restrictive in large dimensions, 
	reflecting 	the ``curse of dimensionality'' that appears frequently in the pattern recognition 
	literature.
	In addition to (\ref{cond_card_1}), we also need to bound the amount of support vectors that
	the solution for the soft margin $L^1$ problem can have in a small cube. Assume that, 
	when the sample size is $n$, $\cal S$ is covered (by weak grid compatibility),
	with a regular array ${\cal W}_n$ such that each cube $V\in {\cal W}_n$ has sides of length
	\[
	l=l_n=\kappa(\logn^2 n /n)^{1/d}
	\]	
	for some positive constant $\kappa$. We assume that there exists a constant $C_1$
	\beq\label{cond_card2}
	\Pr(\max\,\{\#\left({\cal V}_n\cap V\right): V\in {\cal W}\,\}> C_1)\rightarrow 0, 
	\>\>\mbox{ as }\>\>n\rightarrow \infty,
	\enq 
	where, again, ${\cal V}_n$ is the set of support vectors for the solution of the SVM problem.
	Although neither of the two last conditions implies the other, (\ref{cond_card_1}) is, by far, 
	more
	restrictive than (\ref{cond_card2}) on the total number of support vectors that the problem
	might admit, since the second condition does not reflect the curse of dimensionality. 
	Condition (\ref{cond_card2}) imposes a sublinear limit on the number of
	support vectors that can appear in any particular small neighborhood.

	As a first result we prove that for a subsample of size $\lceil \delta n\rceil$, with high 
	probability, there will exist in the subsample, observations  with labels of both classes, close 
	to the support vectors of the complete sample solution. Since the support vectors delimit the 
	surface that separates the two classes, it is natural to expect that in a neighborhood of each 
	support vector, points from both classes are to be found, even in a subsample.
	
	\begin{proposition}\label{covering}
		
	Under the setting described above, there exists a $\rho=\rho(n)$, which is of the order 
	$\mathrm{O}((\logn^2 n /n)^{1/d})$ such that for each $X_{i} \in {\cal V}_{n}$ there are $X_{i'} \in \cal{M}$ with $Y_{i'}=1$ and $X_{i''}\in \cal{M}$ with $Y_{i''}=-1$ such that
		
		\begin{displaymath}
		\|X_{i} - X_{i'} \| \leq \rho \quad \text{y} \quad \|X_{i}-X_{i''}\|\leq \rho
		\end{displaymath}
	\end{proposition}
	
	\emph{Proof:} 
	From the assumptions, there are positive constants $a_{1}$ and $a_{2}$, such that 
	$0<a_{1}\leq f(x)\leq a_{2}, \forall x \in \cal S$ and $b_{1}$ and $b_{2}$, such that 
	$0 < b_{1} \leq \eta(x) \leq b_{2} < 1$. 
	By the grid compatibility assumption, there exists a cover  of $\cal S$, $\cal W$,
	composed of cubes $E_{n,j}$ of side $\ogd\left(\left(\frac{\logn^2 n}{n}\right)^{\frac{1}{d}}
	\right)$. 
	The volume of each cube is of the order $\ogd\left(\frac{\logn^2 n}{n}\right)$.

	By weak grid compatibility, there exists a positive $\gamma$ such that, for each $j$
	
	\begin{equation}\label{vol}
	\mu(E_{n,j}) = \mu(E_{n,j} \cap {\cal S}) \geq \gamma a_{1} \frac{\logn^2 n}{n}. 
	\end{equation}
	
	Let 
	\begin{displaymath}
	B_{n,j}= E_{n,j} \times \{1\}\qquad\mbox{and}\qquad
	{\cal N}^{+}_{n,j}= \# \{(X_{\ell},y_{\ell}) \in B_{n,j}\cap {\cal M}\}
	\end{displaymath}	
	
	\noindent ${\cal N}^{+}_{n,j}$ is the number of observations of the subsample $\cal M$, in 
	$E_{n,j}$, whose class is 1. Also, define
	
	\begin{displaymath}
	p_{n,j}= \prob((X,y)\in B_{n,j}),
	\end{displaymath}
	where $(X,y)$ is a new pair produced by the same random mechanism generating the sample.
	The distribution of  ${\cal N}^{+}_{n,j}$ is binomial with parameters $n'=\lceil \delta n 
	\rceil$ and $p_{n,j}$ i.e. ${\cal N}^{+}_{n,j} \sim \text{Bin}(n',p_{n,j})$ and, by
	our assumptions, 
	$\gamma a_{1}b_{1}\logn^2 n/n \leq p_{n,j} \leq \gamma a_{2}b_{2}\logn^2 n/n$.
	Then, the probability of finding no subsample observations in $E_{n,j}$, for a fixed $j$, is
	
	\begin{displaymath}
	\begin{aligned}
	\prob \Big( {\cal N}^{+}_{n,j} = 0 \Big) & \leq \text{Bin}(n',p_{n,j})(0) \\
	&  \leq \expn \Bigg\{ -\delta  \gamma a_{1}b_{1}{\logn^2 n} \Bigg\}\\
	&= n^{-  \delta \gamma a_{1}b_{1}\logn n }.
	\end{aligned}	  
	\end{displaymath}

	\noindent Notice that by (\ref{vol}), we have
	
	\begin{equation}\label{cardi}
	\# {\cal W} \leq \frac{n}{\gamma a_{1} \logn^2 n},
	\end{equation}

	\noindent It follows that the probability of having no points of the subsample of class 1 in
	some of the cubes in $\cal W$ can be bounded as

	\beqa\label{subsample_bound}
	\prob \Big( \cup_{j} \{{\cal N}^{+}_{n,j} = 0 \}\Big) & \leq  \frac{n}{\gamma a_{1} \logn^2 n} 
	\prob \Big( {\cal N}^{+}_{n,j} = 0 \Big)\nonumber\\  	
	& \leq \frac{n}{\gamma a_{1} \logn^2 n} \cdot  n^{-  \delta \gamma a_{1}b_{1}\logn n }
	\nonumber\\
	& = \frac{1}{\gamma a_{1} \logn^2 n} n^{1-  \delta \gamma a_{1}b_{1}\logn n}.
	\enqa
	
	Since the bound on the right hand side of (\ref{subsample_bound}) summed over $n$, adds to a 
	finite value, we get, by Borel-Cantelli, that the event: ``There exists a cube in $\cal W$
	that contains no points of the subsample of class 1'' will not occur, for large enough $n$,
	almost surely. In other words 
	\beq\label{Borel-Cantelli}
	\prob \Big( \limsup_n\cup_{j} \{{\cal N}^{+}_{n,j} = 0 \}\Big) = 0
	\enq
		
	The version of (\ref{Borel-Cantelli}) for subsample points with $y_i=-1$ is obtained similarly. 
	Since every support vector in ${\cal V}_{n}$ must fall in a $E_{n,j}$, for some $j$,
	the result follows.
	\qeda
	
	The assumptions in Proposition \ref{covering} may seen too restrictive, in asking that both
	$f$ and $\eta$ be bounded away from extreme values on the whole domain $\cal S$. Those
	requirements could be weakened, by requiring those conditions to hold only in regions of 
    $\cal S$ where support vectors might appear. Then, the argument in the proof
    would not consider all cubes in $\cal W$, but only the collection 
   ${\cal E}=\{ E_{n,j}\in {\cal W}:{\cal V}_{n}\cap E_{n,j} \neq \emptyset \}$, that is,
   those cubes in the grid that contain support vectors.
	
\subsection{Stability of the SVM solution}
	
The following theorem is stated in the context of the soft margin $L^1$ SVM problem.
The corresponding result for the soft margin $L^2$ version of the problem also holds, and
is easier to prove in that case.

	\begin{theorem}
		Fix $\epsilon$ greater than 0. 
		Let $\alpha^{*}_{i}$ and $b^*$ be the multipliers appearing in (\ref{class}) 
		for the solution of the SVM problem associated to the complete training data set.  There
		exists a constant $M\geq 1$, such that
		\neno (i) If we replace $\alpha^{*}_{i}$ and $b^*$ 	by $\gamma^*_i=\alpha^{*}_{i}/M$ and 
		$c^*=b^*/M$ in (\ref{class}), the new set of coefficients
		continue to be feasible for the problem (\ref{sml1dual}) and the two corresponding classifiers
		(with the original coefficients and the coefficients divided by $M$)
		coincide, that is, produce always the same classification on new data points.
		\neno (ii) For the soft margin $L^1$ problem on the subsample ${\cal M}$, there exist 
		multipliers $\beta^*_{\ell}$, feasible for the problem (\ref{sml1dual}) on  ${\cal M}$, such
		that, with high probability,
		
		\begin{equation}\label{approx}
		\Big |\sum_{i}y_{i}\gamma^{*}_{i}K(X_{i},x) - \sum_{\ell} y_{\ell}\beta^*_{\ell}K(X_{\ell},x)
		\Big | < \epsilon \qquad\mbox{for all }\> x\in {\cal S},
		\end{equation}
		where the $(X_{\ell},y_{\ell})$ are the data points in $\cal M$.
		\neno (iii) Let class$_{\,\mbox{full}}(x)$ be the classifier defined by (\ref{class}) and
		obtained from the solution of the $L^1$ soft margin SVM problem for the complete sample
		and class$_{\mbox{sub}}(x)$ the classifier defined by	
	\beq\label{class2}
	\mbox{class}_{\mbox{sub}}(x)=\mbox{sgn}\left(\sum_{\ell}y_{\ell}\beta^*_{\ell}K(X_{\ell},x)+c^*\right)
	\enq 
Then, with high probability,
\beq\label{iii}
	\mu\left(x\in{\cal S}:\mbox{class}_{\,\mbox{full}}(x)\neq \mbox{class}_{\mbox{sub}}(x)\right)
	<\epsilon
	\enq
	
	\end{theorem} 
	
	\emph{Proof:} 
	
	Let $B$ be the set of probability 1 where (\ref{Borel-Cantelli}) holds. We assume that
	our sample, \neno $(X_1,y_1),(X_2,y_2),\dots$ falls in that set. Consider the cube array, 
	$\cal W$, of
	the previous proposition and the subcollection
	${\cal E}=\{ E_{n,j}\in {\cal W}:{\cal V}_{n}\cap E_{n,j} \neq \emptyset \}$ of cubes where
	the solution of the SVM problem for the complete sample of size $n$ has support vectors.
	Let $j$ be such that $E_{n,j}\in {\cal E}$. Define 
	\[
	\begin{aligned}
	& A^+_{n,j}=\{i\leq n: \alpha^*_i>0,X_i\in E_{n,j}\mbox{ and }y_i=+1\}, \\
  & A^-_{n,j}=\{i\leq n: \alpha^*_i>0,X_i\in E_{n,j}\mbox{ and }y_i=-1\},	
	\end{aligned}	
	\]
	and let
	\[
	s^+_{n,j}=\sum_{i\in A^+_{n,j}}\alpha^*_i\qquad\mbox{ and }\qquad
	s^-_{n,j}=\sum_{i\in A^-_{n,j}}\alpha^*_i
	\]
	in the understanding that a sum over the empty set is 0. $s^+_{n,j}$ is the sum of
	the multipliers associated to support vectors with class +1 in the cube $E_{n,j}$.
	We want to assign the sum of multipliers in the cube
	as multiplier for a vector of the subsample
	in $E_{n,j}$. But then, the new set of coefficients could fail to satisfy the upper bound
	restriction $\alpha_i\leq C$ in (\ref{sml1dual}). For that reason, we consider the sets 
	$S^+=\{s^+_{n,j}/C, j\geq 1\}$ and $S^-=\{s^-_{n,j}/C, j\geq 1\}$ and 
	$M=\max\left(S^+\bigcup S^-\bigcup \{1\}\right)$.
	All the $s^+_{n,j}$ and $s^-_{n,j}$, divided by $M$ are bounded above by $C$.
	It is simple to verify that, by defining $\gamma^*_i=\alpha^{*}_{i}/M$, 
	for each support vector in ${\cal V}_{n}$
	and $c^*=b^*/M$, we obtain a new set
	of multipliers that satisfies the feasibility conditions (\ref{sml1dual}) and that produce
	the same classification for each $x\in\cal S$ as the original classifier. Thus, 
	part (i) is proved. As a bonus, we observe that, by assumption (\ref{cond_card2}), $M\leq C_1$
	for the value of $C_1$ appearing in that assumption.
	
	Since we are working in $B$ and the sample is large enough, for each $j$ such
	that $E_{n,j}\in{\cal E}$, there exists at least one $X_\ell\in {\cal M}\cap E_{n,j}$ with
	$y_{\ell}=+1$. Choose one of those $X_\ell$, and assign to it the multiplier 
	\beq\label{betaell}
	\beta^*_{\ell}=\sum_{i\in A^+_{n,j}}\gamma^*_i
	\enq
	Define similarly $\beta^*_{\ell}$ for a point $X_\ell\in {\cal M}\cap E_{n,j}$ with
	$y_{\ell}=-1$ in terms of the $\gamma^*_i$ for support vectors in $E_{n,j}$ with $y_i=-1$.
	For the $X_\ell$ not chosen, set $\beta_{\ell}=0$.
	By construction, the set of coefficients $\beta^*_{\ell}$ just defined, together with
	$c^*=b^*/M$ is feasible as solution associated to $\cal M$. 
	
	Now, for each $x\in\cal S$, and the chosen $X_\ell\in {\cal M}\cap E_{n,j}$ with
	$y_{\ell}=+1$, we have, using (\ref{betaell}), that $K$ is Lipschitz and that the
	diameter of $E_{n,j}$ is $\mathrm{O}((\logn^2 n /n)^{1/d})$ by the 
	construction of the previous Proposition,
	\beqa\label{Lipschitz}
	&|\sum_{i\in A^+_{n,j}}y_i\gamma^*_iK(X_i,x)-y_{\ell}\beta^*_{\ell}K(X_{\ell},x)|=
	|\sum_{i\in A^+_{n,j}}\gamma^*_i(K(X_i,x)-K(X_{\ell},x))|\leq \nonumber \\
	&\# A^+_{n,j}\,C\,L\,\mathrm{O}((\logn^2 n /n)^{1/d})=
	\# A^+_{n,j}\,\mathrm{O}((\logn^2 n /n)^{1/d}).
	\enqa
	Adding over all sets $A^+_{n,j}$ and $A^-_{n,j}$, we get 
\[
	\Big |\sum_{X_i \in {\cal V}_{n}}y_{i}\gamma^{*}_{i}K(X_{i},x) - \sum_{\ell} y_{\ell}		
	\beta^*_{\ell} K(X_{\ell},x)\Big | \leq \# {\cal V}_{n}\mathrm{O}((\logn^2 n /n)^{1/d})
\]
which goes to zero, in probability, by our assumption on $\# {\cal V}_{n}$. This ends the
proof of (ii).

	By part (i), for the event in (\ref{iii}) to occur, the classification produced by 
	$\mbox{class}_{\mbox{sub}}(x)$ and the classifier 
	$\mbox{class}_{2}(x)=\mbox{sgn}\left(\sum_{i}y_i\gamma_i^*K(X_i,x)+c^*\right)$
	must differ. Suppose that 
	\[\left|\sum_{\ell}y_{\ell}\beta^*_{\ell}K(X_{\ell},x)+c^*\right|>
	\epsilon\]. 
	Then, for the classifiers to differ in their decisions, we must have
	\[
	\Big |\sum_{i}y_{i}\gamma^{*}_{i}K(X_{i},x) - \sum_{\ell} y_{\ell}\beta^*_{\ell}K(X_{\ell},x)
	\Big | > \epsilon, 
	\]
	an event of probability that goes to 0, by (\ref{approx}). Suppose now that the classifiers
	differ when
	\[
	\left|\sum_{\ell}y_{\ell}\beta^*_{\ell}K(X_{\ell},x)+c^*\right|\leq
	\epsilon. 
	\] Then, using part (ii), we can assume, with high probability, that
	$\left|\sum_{i}y_i\gamma_i^*K(X_i,x)+c^*\right|\leq 2\eps$. Multiplying by $M$, and
	using the observation on the value of $M$ made at the end of the proof of (i), it follows that
	\beq\label{cond_on_class_full}
	\left|\sum_{i}y_i\alpha_i^*K(X_i,x)+b^*\right|\leq 2C_1\eps
	\enq
	for the value of $C_1$ in (\ref{cond_card2}). The probability $\mu(\cdot)$ of the set
	of $x\in \cal S$ satisfying (\ref{cond_on_class_full}) goes to 0, as $n$ grows, 
	by the assumption on $\epsilon$-uncertainty, (\ref{asympcont}), finishing the
	proof. 	$\qeda$

	\subsection{Numerical evaluation of probability of $\epsilon$-indecision}\label{indecision}
	
	One of the key assumptions made to obtain the results in the previous subsection is condition 
	(\ref{asympcont}), which says that, when $\epsilon$ goes to 0 and $n$ is large, the
	probability of the region of $\epsilon$-indecision of the classifier produced by the
	solution of the SVM problem, goes to zero as well.
	
	On a particular data set, the probability of the event in (\ref{asympcont}) can be approximated
	as the fraction of $x$'s in the sample, for which the condition
	$\left|\sum_{i}y_i\alpha_i^*K(X_i,x)+b^*\right|\leq\epsilon$ holds, for different values 
	of $\epsilon$ and sample sizes.
	
	Such a numerical evaluation is described next. The problems considered are COD-RNA, IJCNN1,
	and WEBSPAM from the LibSVM library \footnote{See Section 3 for a description of the datasets.}. 
	
	For computational reasons, due to the size of the simulations required, in the experiment to be described next, only the linear kernel was used. Tables \ref{table:margencodrna}, 
	\ref{table:margenijcnn} and \ref{table:margenwebspam} present averages of the 
	estimated empirical fraction
	\beq\label{empirical_indecision}
	\frac{\# \{j\leq n : \Big |\sum_{i}y_{i}\alpha^{*}_{i}K(X^{*}_{i},X_j)+b^{*} \Big | < \epsilon 
	\Big \}}{n}
	\enq
	computed over the training set as follows. Each $X_j$ of the training subsample is
	included in the calculation, the $X_{i}^{*}$ are the support vectors for that training subsample
	and $n$ is the subsample size.  From the data available for training in the data set,
	random subsamples of size $n$, for different choices of $n$ are taken, and the corresponding
	SVM problems are solved and the fraction (\ref{empirical_indecision}) estimated. For each training sample size considered, the experiment is repeated 100 times, for independent 
	for independent subsamples and the
	average of the empirical indecision probabilities is reported in the tables. 
	
	\begin{table}[!ht]
		\caption{Empirically estimated indecision probability. COD-RNA data} 
		\label{table:margencodrna}
		\centering
		\begin{tabular}{|l |c| c| c| c| c| c|}
			\hline\hline		
			\multicolumn{2}{|c}{} Linear kernel  &  &  & \\
			\hline
			\hline
			$\epsilon$	&  $n$ = 5000 & $n$ =10000 & $n$ =20000 & $n$=30000 & $n$=40000 & $n$=50000 \\
			[0.5ex]
			\hline
			0.1 & 0.07758 & 0.07726   & 0.07773 &  0.07754 & 0.07759 & 0.07751 \\
			0.01 &  0.00722  &  0.00722  & 0.00726 & 0.00724  & 0.00733 & 0.00729 \\
			0.001 &  0.00082 &  0.00077 & 0.00078 &  0.00078 & 0.00079 & 0.00078 \\
			\hline
		\end{tabular}
	\end{table}

	\begin{table}[!ht]
		\caption{Empirically estimated indecision probability. IJCNN1 data} 
		\label{table:margenijcnn}
		\centering
		\begin{tabular}{|l |c| c| c| c| c| c|}
			\hline\hline		
			\multicolumn{2}{|c}{} Linear kernel  &  &  & \\
			\hline
			\hline
			$\epsilon$	&  $n$ = 5000 & $n$ =10000 & $n$ =20000 & $n$=30000 & $n$=40000 & $n$=49990 \\
			[0.5ex]
			\hline
			0.1 & 0.15623 & 0.15611   & 0.15675 &  0.15673 & 0.15655 & 0.1565 \\
			0.01 &  0.01147  &  0.01141  & 0.01148 & 0.01144  & 0.01141 & 0.01142 \\
			0.001 &  0.00086 &  0.00082 & 0.00085 &  0.00086 & 0.00085 & 0.00086 \\
			\hline
		\end{tabular}
	\end{table}

	\begin{table}[!ht]
		\caption{Empirically estimated indecision probability. WEBSPAM  data}
		\label{table:margenwebspam}
		\centering
		\begin{tabular}{|l| c| c| c| c| c| c|}
			\hline\hline		
			\multicolumn{2}{|c}{} Linear kernel  &  &  & \\
			\hline
			\hline
			$\epsilon$	&  $n$ = 10000 & $n$ =20000 & $n$ =30000 & $n$=60000 & $n$=120000 & $n$=240000 \\
			[0.5ex]
			\hline
			0.1 & 0.99963 & 0.99966   & 0.9996 &  0.99962 & 0.9996 & 0.9996 \\
			0.01 &  0.2391  &  0.23892  & 0.2396 & 0.2389  & 0.23989 & 0.23983 \\
			0.001 &  0.01736 &  0.017 & 0.01716 &  0.017145 & 0.017 & 0.01724 \\
			\hline
		\end{tabular}
	\end{table}
	
From the numbers in these tables, it appears that in all three examples considered, 
for each value of $\epsilon$, the average estimated indecision probability is converging
to a limiting value as $n$ grows and, the indecision probability decreases with $\epsilon$,
in a nearly linear fashion. We conclude that assumption (\ref{asympcont}) seems to be
valid in diverse real examples.

	\section{Conclusions}
	
	A local bagging methodology for SVM classification has been proposed. It is local in the sense that uses information close to the observations of interest (support vectors).
	The bagging and local subsampling SVM methodology presented here can attain, in many problems, a classification error comparable to that  corresponding to the solution of the problem on the full training sample, using a fraction of the training data set and a smaller number of support
vectors, so a significant saving on computational time is obtained. The advantages of the methodology introduced are more noticeable when the
	dimension of the data set is not very large and depend, to some extent, on the complexity
	of the problem, and on the kernel used in the algorithm.

	\section*{Acknowledgements}
	
	We would like to thank the Institution Universidad Militar Nueva Granada, who supported part of this work through projects INV-CIAS 2545-2544 by author MDGL. Research by authors RB and JO was partially supported by CONACYT, Mexico Project 169175. This work was partly done while RB visited the Departamento de Matematicas, Universidad de los Andes, Colombia (RB as visiting graduate student supported by Mixed Scholarship CONACYT, Mexico). Their hospitality and support is gratefully acknowledged. 
	

	
	
	
\end{document}